\documentclass[11pt,a4paper]{article}
\usepackage[hyperref]{acl2023}
\usepackage{times}
\usepackage{tipa}
\usepackage{latexsym}

\usepackage{graphicx}
\usepackage{float}
\usepackage{subfig}

\usepackage{microtype}
\usepackage{natbib}
\usepackage{hyperref}

\def\aclpaperid{4171} 

\title{Exploring How Generative Adversarial Networks Learn Phonological Representations}

\author{Jingyi Chen \\
  Department of Linguistics\\
  The Ohio State University  \\
   \texttt{chen.9220@osu.edu}\\
   \And
  Micha Elsner \\
  Department of Linguistics  \\
  The Ohio State University \\
  \texttt{elsner.14@osu.edu}\\
   }

\date{}

\begin{document}
\maketitle

\begin{abstract}
This paper explores how Generative Adversarial Networks (GANs) learn representations of phonological phenomena. We analyze how GANs encode contrastive and non-contrastive nasality in French and English vowels by applying the ciwGAN architecture \cite{begus2021ciwgan}. \citeauthor{begus2021ciwgan} claims that ciwGAN encodes linguistically meaningful representations with categorical variables in its latent space and manipulating the latent variables shows an almost one to one corresponding control of the phonological features in ciwGAN's generated outputs. However, our results show an interactive effect of latent variables on the features in the generated outputs, which suggests the learned representations in neural networks are different from the phonological representations proposed by linguists. On the other hand, ciwGAN is able to distinguish contrastive and noncontrastive features in English and French by encoding them differently. Comparing the performance of GANs learning from different languages results in a better understanding of what language specific features contribute to developing language specific phonological representations. We also discuss the role of training data frequencies in phonological feature learning.
\end{abstract}

\section{Introduction}
Recent studies in natural language processing (NLP) have demonstrated two generic trends: neural networks dominate language-specific machine learning models; the common practice of model training (pre-training and fine-tuning) outperforms many traditional training methods and is particularly suitable for the development of language models used for various downstream tasks.
These language models, however, are of black-box nature. The interpretability of these models is limited that the language representation they learned might not align to human language. How, then, to understand the opaque and complex learned representation of language models is an important question in recent studies. Phonology, the study of the sound system of human languages, plays an important role in understanding models' inherent biases and their ability to make human-like generalizations.

The sound systems of human languages are not organized arbitrarily, but contain structural generalizations and interdependence. Thus, learning a sound system involves not only learning to acoustically realize or recognize segments (phonetics), but also mapping them to an inventory characterized by distinctive features, and learning distributional constraints on segment sequences (phonology). Just as computational psycholinguists have investigated the degree to which neural network language models learn linguistically motivated features like syntax \cite{linzen2016assessing,https://doi.org/10.1111/cogs.12414,gulordava-etal-2018-colorless, marvin-linzen-2018-targeted, futrell-etal-2019-neural}, they have also investigated the degree to which phonological organization emerges from neural models trained on acoustics \cite{gelderloos2016phonemes, chrupala-etal-2017-representations}.

The degree to which these models learn phonological features is still debatable.
Recently, a neural network autoencoder seems to successfully learn phoneme-like representations without explicit labels \cite{rasanen2016analyzing, shain2019measuring}. While autoencoders seem to acquire some phonological generalizations, their representations of the kind of phonological features used by linguists are both incomplete and distributed across the latent space, requiring probing classifiers to detect. Because of this limited success and lack of transparency, it is difficult to tell whether higher-order phonotactic dependencies between different segments are acquired.
Generative Adversarial Networks (GANs) \cite{goodfellow2014generative,goodfellow2020generative,begus2020modeling}, on the other hand, are claimed to model language acquisition naturally because GANs can model phonetic and phonological computation as an almost one to one mapping from random space to generated data of a GAN instance trained on raw speech data \cite{begus2022interpreting}. The learned internal representations of GANs is claimed to resemble phonological learning in human speech acquisition: GANs learn to build their internal latent space via unsupervised phonetic learning from raw acoustic data, which is similar to human constructs underlying phonological representation by listening to the speech sounds in a language.

\newcite{begus2021ciwgan} proposed ciwGAN (Categorical InfoGAN) which is based on WaveGAN architecture but with an extra Q-network that motivates the Generator to produce linguistically categorical and meaningful sounds.
\newcite{begus2022interpreting} shows that ciwGAN can encode allophonic distribution: word-initial pre-vocalic aspiration of voiceless stops ([\textipa{p\textsuperscript{h}\textipa{I}t}] v.s. [sp\textipa{I}t]). In English, the aspiration of stop consonant T occurs initially before a vowel (\#T\textsuperscript{h}V, \textsuperscript{h} refers to the aspiration) while a period of stop closure occurs between the aspiration and the period frication noise of [s] (\#sTV). CiwGAN successfully learned and generated this allophonic distribution in that the generated outputs obey this phonological constraint. Moreover, changing a single variable in the latent space is capable of changing generated tokens from sTV to T\textsuperscript{h}V, suggesting an almost one-to-one correspondence between latent variables and phonological features. This finding is claimed to prove that GANs can model unsupervised phonological representation learning from raw speech data.

In this study, we explore the robustness of ciwGAN as a phonological feature learner by testing ciwGAN on learning the feature of nasality, which is distinct in French and English. Nasality is a contrastive feature for French vowels; nasal vowels can appear independently of nasal consonants \cite{cohn1993nasalisation}. In English, however, vowel nasality is allophonic, like voiceless stop aspiration -- nasal vowels appear only preceding nasal consonants. Linguists traditionally analyze this relationship as reflecting a single nasal feature on the consonant, without an independent feature controlling vowel nasality
\cite{kager1999optimality, mcmahon2002introduction, hayes2011introductory, ogden2017introduction, zsiga2012sounds}. Thus, our experiment provides a more rigorously controlled test of the claims of \newcite{begus2022interpreting}.
CiwGAN networks are trained on English and French datasets respectively to learn the distinct nasal features of the two languages. Analysis of the result ciwGAN networks is development to answer the following research questions: (1) What features of the data contribute to learning the nasal representations in English vs. French? (2) How does the training data’s distribution affect the learned feature system in waveGAN network?

Results show interactive effects between latent variables in controlling the phonetic and phonological features: multiple to one corresponding mapping is found between latent variables and the phonetic and phonological features, suggesting that the claimed advantage of GANs over autoencoders is not as great as was originally claimed. ciwGAN do react differently in encoding the different nasal representations in English and French to indicate whether a feature is or is not contrastive, highlighting their potential as phonological learners. Moreover, we found that training data’s distribution affects the learned feature system in ciwGAN; to the extent that GANs can be considered cognitively plausible models of human learning, this may lead to predictions about how changes in phonetic distribution can become phonologized into almost-categorical rules.




\section{Related Works}
We review two areas of recent literature. Large-scale unsupervised models of speech learn words and in some cases phoneme categories, but the degree to which they acquire phonological feature systems is not clear. Some smaller-scale models have been specifically analyzed in phonological terms.
One recent and successful pre-trained model (wav2vec 2.0) is shown to encode audio sequences with its intermediate representation vectors, which demonstrates superiority in downstream fine-tuning such as automatic speech recognition (ASR) tasks,  speaker verification tasks, and keyword spotting tasks \cite{baevski2020wav2vec}.

Similar to wav2vec, Hu-BERT \cite{hsu2021hubert}, a pretrain language model that leverages self-supervised learning for speech, directly processes audio waveform information from raw speech to predict clustering categories for the speech segments. Both wav2vec 2.0 and Hu-BERT have been successful in capturing acoustic information from raw speech and improve the state-of-the-art performance in speech recognition and translation.
\newcite{oord2016wavenet} introduces a dilated causal convolutional network WaveNet which attempts to discover phone units from audios; however, because of the lack of lexical knowledge, WaveNet cannot emit explicit phonemes \cite{oord2016wavenet}.

Moreover, the submissions for the ZeroSpeech Challenges \cite{dunbar2017zero, dunbar2019zero, dunbar2020zero, dunbar2021zero} utilizes generative models like  GANs \cite{begus2021ciwgan,yamamoto2020parallel} and autoencoders \cite{chung2016audio,baevski2020vqwav2vec} to learn the lexical or phone-level presentation from raw speech data. However, the learning of phonology features of language from raw speech data is not particularly implemented or evaluated in the above studies.
Although these models have shown impressive results in speech representation learning that capture phonetic/acoustic content, the degree to which they acquire phonological feature systems is still not clear.

Some studies have been focused on developing language models that learn phonological representations. In \citet{shain2019measuring}, an autoencoder neural network is trained on pre-segmented acoustic data and output values that correlates to phonological features.
Nevertheless, the architecture of autoencoder brings a problem in learning phonological representation: because autoencoders are trained to reproduce their inputs faithfully, their latent representations may contain too much information which is extraneous to phonological categorization, such as speaker-specific information. GANs are not trained to strictly reproduce the training data and therefore might not be subject to this issue.

Recently, \newcite{donahue2019impact}'s study applies the GAN architecture based on the DCGAN architecture \cite{radford2015unsupervised} to learn language features from
continuous speech signals (WaveGAN). GAN networks as generative model, is firstly applied in learning allophonic distribution from raw acoustic data in \citet{begus2020generative,begus2020modeling} which also proposes a probing technique to interpret the internal representation of GAN networks. 
The internal language representation is probed and claimed to be interpretable in \newcite{begus2021identitybased,begus2022interpreting} which  firstly shows that GAN networks can learn reduplication and conditional allophonic distribution of voice onset time (VOT) duration from the raw speech audio, respectively. 

\newcite{begus2021ciwgan} proposes ciwGAN (Categorical InfoWaveGAN) and fiwGAN, two GAN networks for unsupervised lexical learning from raw acoustic inputs; the two GAN networks combine WaveGAN with InfoGAN, an extension to GAN architecture, that includes an additional ``Q-network'' which encourages the model’s productions to group into discrete categories \cite{chen2016infogan}. In these earlier papers, the discrete representational elements in these GAN architectures were proposed and interpreted with respect to lexical category learning. In our work, this interpretation does not apply, since our data consists of syllables rather than whole words. While top-down lexical information appears critical to learning many phonological contrasts, the rules governing the distribution of vowel nasality we are studying here are local phonotactic phenomena which can be learned purely by capturing the distribution of vowels and coda consonants.
\section{Model}

\begin{figure}[ht]
	\centering
	\includegraphics[width=1\linewidth]{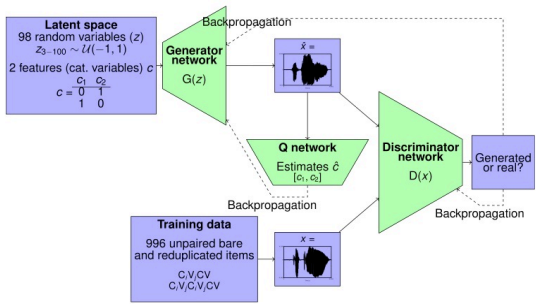}
	\caption{ciwGAN architecture: three convolutional neural networks are presented by green boxes and inputs to these neural networks are presented by purple boxes. This figure is from \cite{begus2022interpreting}.}
	\label{fig:ciwGAN}
\end{figure}

In this paper, we use ciwGAN to model phonetic and phonological learning for vowel nasalization in English and French. 
The GAN architecture involves two deep convolutional neural networks: the Generator network and the Discriminator network \cite{goodfellow2014generative,goodfellow2020generative}. They are trained against each other to boost their performance. The Generator network is trained to generate data from a set of latent variables and maximize the error rate of the Discriminator network. The Discriminator takes the training data and output of the Generator network as input and attempts to determine whether its input comes from the training dataset (actual data) or generator output (fake data). The competition of the two networks against each other makes the Generator generate data that is similar to the actual data.
The architecture of ciwGAN is shown in \autoref{fig:ciwGAN}. The Generator takes categorical binary latent variables $\phi$ (size is 3 in \autoref{fig:ciwGAN}) and continuous latent variable $z$ that are uniformly distributed in the interval (-1, 1) as input and outputs a continuous time-series data as audio signal ($\hat{x}$). The Q-network, extra component in ciwGAN than WaveGAN, also takes audio signals as input, but gives a categorical estimation $\hat{\phi}$ on the audio signal. It is trained to minimize the difference between the categorical estimation  $\hat{\phi}$ and the actual latent categorical variables $\phi$ in the Generator's latent space.
With the Q-network, the Generator is motivated to generate audio signals that are categorically distinguishable for the Q-network.



To interpret the learned phonological features in the generated output, \citet{begus2022interpreting} uses regression analysis. They manually label each generated audio snippet with its phonological features, then measures the strength of correlation between the latent variables ($z$) and the phonological feature of interest. We also use this technique in our experiments to find the latent variables that correspond to the nasal feature in English and French.
\citet{beguvs2020generative} uses regression analysis from the latent variables to the phonetic and phonological features in the generated outputs to reveal the correspondence relations between latent variables and the phonetic and phonological features.

However, to avoid expensive manual labeling, we develop a supervised nasal detector (nasalDNN), a deep neural network model adapted from \citet{yurt2021fricative}, to determine whether a generated output carries nasality or not. The nasalDNN is a 1D CNN that takes speech segments as inputs, and calculates the posterior probabilities for the sample at the center point of the segment belongs to nasal phoneme classes [n, m, ng].

For French, we trained the convolutional nasalDNN on the SIWIS dataset, which has ground truth labels for both nasal consonants and nasal vowels. We used these labels to learn a four-way classifier, which we applied to the sample at the center point of each segment. In English, since TIMIT has no ground truth labeling of nasal vowels, we used a different procedure: we learned independent classifiers for vowels and nasal sounds (using consonants as the gold examples of nasals) and detected nasal vowels by intersecting the predictions.

\section{Data}
\label{sec:data}

To learn vowel and nasality features in Engish and French, two ciwGAN instances are trained separately on TIMIT Speech Corpus \cite{garofolo1993darpa} and the SIWIS French Speech Synthesis Database \cite{yamagishi2017siwis}. The TIMIT Speech Corpus includes English raw speech sentences (at 16 kHz sampling rate) and their corresponding time-aligned phonetic labels. In the TIMIT corpus, there are 6300 sentences recorded by 630 speakers from eight dialect regions of the United States. We used the entire TIMIT dataset to extract training data for the English experiment. The SIWIS French Speech Synthesis Database consists of high-quality French speech recordings and associated text files. There are 9750 utterances uttered by French speakers. This French database includes more than ten hours of speech data.

\subsection{Data Preprocessing}
For English dataset, we first excluded SA sentences in TIMIT, which are read by all the speakers, to avoid a possible bias and then extracted sliced sequences of the structure VT and VN from the rest of the sentences \footnote{T refers to voiced and voiceless stop consonants as well as the stop closures [t, d, p, b, k, g, tcl, dcl, pcl, bcl, kcl, gcl], N refers to three nasal consonants in English [n, m, ng], and V includes vowels and approximants  [aa, ae, ah, ao, ax, ax-h, axr, ay, aw, eh, el, er, ey, ih, ix, iy, ow, oy, uh, uw, ux, r, l, w]}. 6255 tokens are extracted from the monosyllabic words and 2474 are extracted from the multi-syllabic words' last syllable . Thus, altogether 8729 tokens from TIMIT were used for training, 5570 tokens of the structure VT, 3159 tokens of the structure VN.

As the SIWIS French Speech Synthesis Database does not provide time-aligned phonetic labels for their recordings, we use the Montreal Forced Aligner \cite{mcauliffe2017montreal}, a forced alignment system with acoustic models using Kaldi speech recognition toolkit \cite{povey2011kaldi} to time-align a transcript corresponding to a audio file at the phone and word levels. 
Based on the time-aligned phonetic labels, we extracted sliced sequences of the structure VT, VN, \H{V}T, \H{V}N \footnote{The T class is [t, d, p, b, k, g, tcl, dcl, pcl, bcl, kcl, gcl] while N includes [n, m, ng, nj].}. As French has contrastive nasal vowels and oral vowels, we used \H{V} to indicate nasal vowels \footnote{Nasal vowels: [\H{A}, \H{E}, \H{o}, \H{OE}] corresponding ipa symbols: [\H{a}, \H{\textipa{E}}, \H{o}, \H{\textipa{oe}} ]} and used V to show oral vowels \footnote{Oral vowels: [A, i, O, AX, a, o, e, u, OE, EU, E] corresponding ipa symbols [a, i, \textipa{OI}, \textipa{@}, o, e, u, \textipa{oe}, \o, \textipa{E}]}. We extracted 4686 tokens where 2681 tokens are extracted from monosyllabic words and 2005 tokens are from the multisyllabic words' last syllable. We have 1031 \H{V}T tokens, 2577 VT tokens, 47 \H{V}N tokens, and 1031 VN tokens as French training dataset. Example lexical items of English and
French are shown in the appendix.
\begin{table}
    \centering
    \begin{tabular}{|c|c|c|c|c|}
        \hline
        Dataset         & VT   & VN   & \H{V}T & \H{V}N \\\hline
        TIMIT$^{\star}$ & 5570 & 3159 & 0      & 0      \\\hline
        SIWIS$^{\star}$ & 2577 & 1031 & 1031   & 47     \\\hline
    \end{tabular}
    \caption{Training Dataset for CiwGAN to Learn Vowel and Nasality Features in English and French}
    \label{tab:_04_trainingdata}
\end{table}

\label{sec:length}

\begin{figure*}[htbp]
	\centering
	\subfloat[English - $z$90 \& $z$13]{\includegraphics[width=.47\linewidth,trim=50 34 50 50,clip]{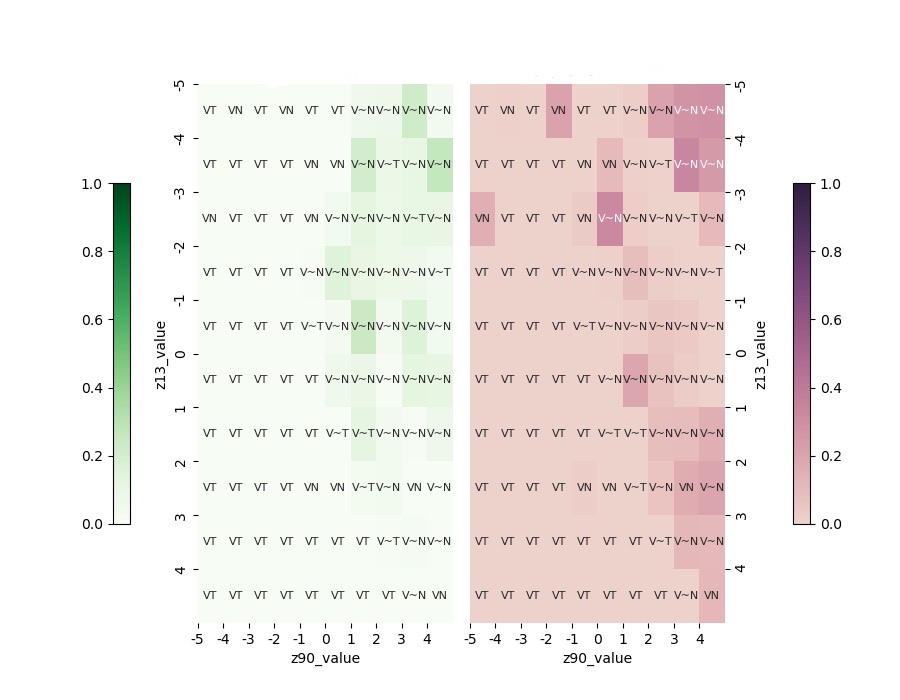}\label{fig:z13}}
	\hfill
	\subfloat[English - $z$4 \& $z$37]{\includegraphics[width=.47\linewidth,trim=50 34 50 50,clip]{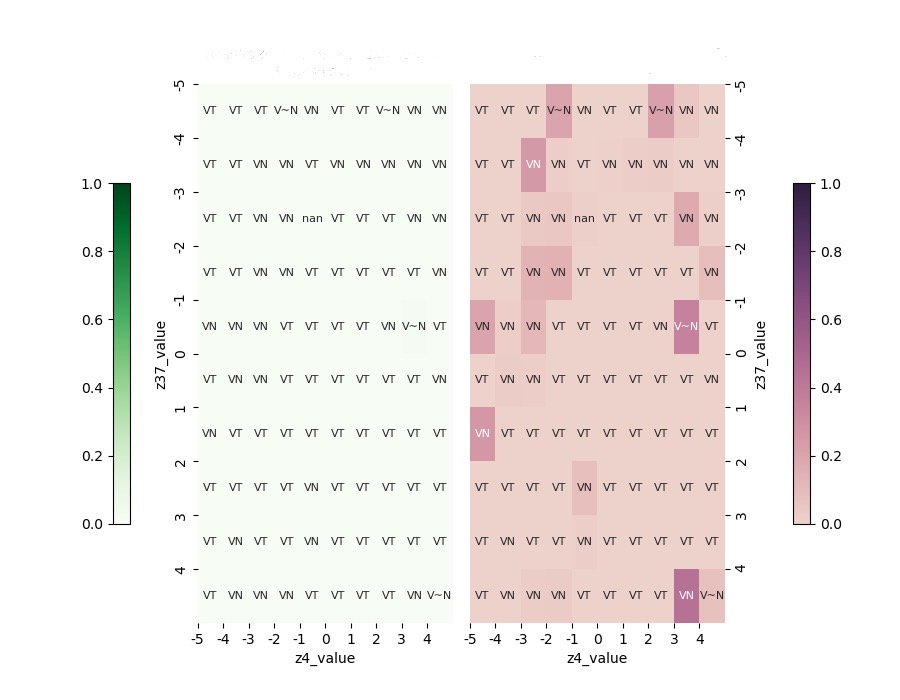}\label{fig:z4}}
    \vspace{-0.2cm}
	\caption{English experiment results: \autoref{fig:z13} shows the generated audios with concurrent manipulation of latent variables $z$90 and $z$13 (x axis: $z$90 \& y axis: $z$13); \autoref{fig:z4} shows the generated audios with concurrent manipulation of latent variables $z$4 and $z$37 (x axis: $z$4 \& y axis: $z$37); Green color heatmap (left side of \autoref{fig:z13} and \autoref{fig:z4}) indicates the detected English nasal vowel on generated audio; Red color heatmap (right side of \autoref{fig:z13} and \autoref{fig:z4}) indicates the detected English nasal consonant on the same generated audio; darkness of color refers to the proportion of detected nasal vowels and the detected nasal consonants in the manipulated audios; annotation are syllables types of the manipulated audios based on the results of nasal detectors.}
    \vspace{-0.5cm}
\end{figure*}
\section{Experiments}
\label{sec:experiment}


To explore our first research question: What features of the data contribute to learning the nasal representations in English vs. French, we implement English and French experiments. The results suggest different learned phonetic/phonological representations in ciwGAN may be caused by different typology of English and French syllable types for nasal vowels and nasal consonants.

\subsection{English Experiment}
\label{ssec:eng_experiment}

After the ciwGAN instance is trained for 649 epochs, it learns to generate 3840 speech-like sequences (VT and VN) that are similar to the training data.
As described above, we label these outputs with a supervised classifier to determine which ones are nasal, then apply linear regression analysis to identify latent variables that correlate to nasal features. The results of linear regression are shown in \autoref{fig:chi} in Appendix. Among the 100 latent variables in latent space, we identify 7 latent variables that have the highest chi-square scores, which indicates a strongly correlation to nasality. \autoref{fig:chi} also illustrates a considerable difference between the highest seven latent variables and the rest of the variables indicating that ciwGAN may encodes nasal feature mainly with these seven latent variables and use other latent variables to increase variance.

We also apply another investigative technique from \newcite{beguvs2020generative}, in which selected latent variables are set to values outside their training range. As in that study, we examine the audio generated from representations with manipulated variables, which contain exaggerated acoustic cues indicating which phonetic qualities the variables control. We sample 100 random latent vectors, and for each one, manipulate the target variable to values between -5 and 5 in increments of 1.
\begin{figure*}[htbp]
	\centering
	\subfloat[French - $z$4 \& $z$37]{\includegraphics[width=.47\linewidth,trim=50 34 50 50,clip]{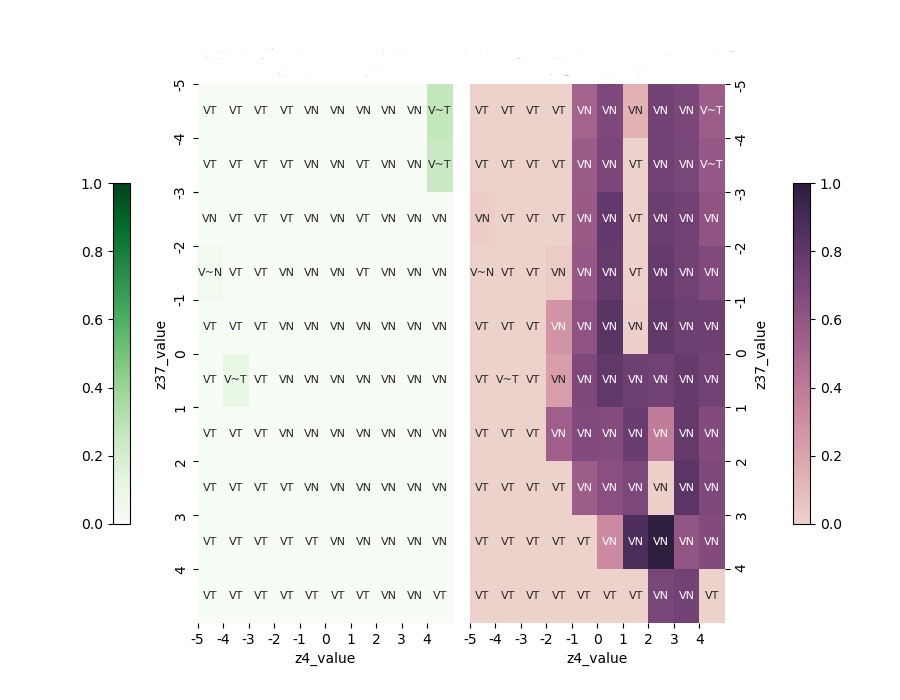}\label{fig:z37}}
	\hfill
	\subfloat[French - $z$88 \& $z$91]{\includegraphics[width=.47\linewidth,trim=50 34 50 50,clip]{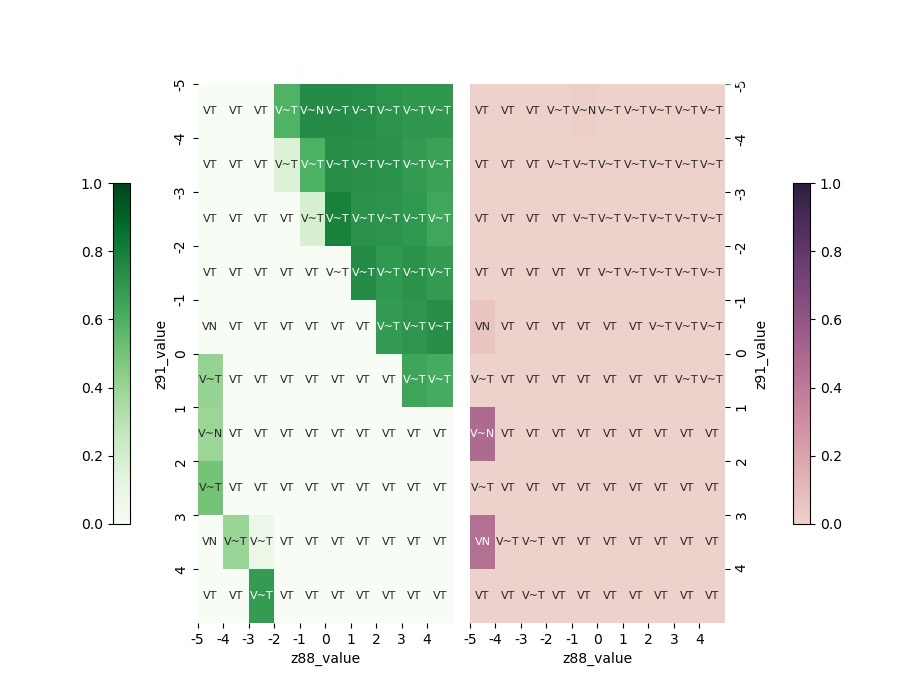}\label{fig:z88}}
    \vspace{-0.2cm}
	\caption{French experiment results: \autoref{fig:z37} shows the generated audios with concurrent manipulation of latent variables $z$4 and $z$37 (x axis: $z$4 \& y axis: $z$37); \autoref{fig:z88} shows the generated audios with concurrent manipulation of latent variables $z$88 and $z$91 (x axis: $z$88 \& y axis: $z$91).  Green heatmap indicates the detected French nasal vowel; Red heatmap indicates the detected French nasal consonant.}
    \vspace{-0.5cm}
\end{figure*}

Although seven latent variables are identified as closely corresponding to the presence of consonants' nasal feature via linear regression, only two latent variables $z$13 and $z$90 show a strong control of the nasality in consonants. \autoref{fig:_05_spectrom_praat_z90z13} , in Appendix, illustrates the manipulation effects of $z$13 and $z$90 on nasal consonant.
The spectrograms show a relatively high F1 (around 650 Hz) initially which corresponds to the vowel and a lower amplitude (F1 at around 250 Hz) at the end of the sound which represents the nasal consonant [n]. The nasality in the consonant gradually decreases as the  values of $z$13 and $z$90 increase separately.
Seven latent variables are also found to be relative to nasal vowels via linear regression; however, manipulating these seven latent variables, vowels' nasality do not show a regular change pattern in the generated audios, which indicates that these seven latent variables do not have one to one corresponding control of the nasality in vowels.

As both latent variables $z$13 and $z$90 are able to control the nasality in consonants, we further explore the interactive effects of these two latent variables by manipulating them simultaneously to test all combinations of the two variables in range [-5,5] and increment of 1. However, no clear interactive correlation are found regarding to the nasality between the two latent variables. Although $z$13 and $z$90 show effects on the nasal feature in consonants when they are manipulated separately, $z$90 show a primary control on consonants' nasality. As illustrated in \autoref{fig:z13}, when $z$90 \textgreater 0, the Generator tends to produce nasal consonants while the value of $z$13 does not show a clear effect on generated sound features. We also found that vowels’ nasality tends to covary with the presence of nasal codas. In \autoref{fig:z13}, whenever a nasal vowel is detected in the generated outputs, they also have a nasal consonant detected in the outputs.

We also evaluate if the two latent variables ($z$4 and $z$37), with the highest chi-square value for nasal vowels, have effects on producing English nasal vowels. However, neither $z$4 nor $z$37 show control of English nasal vowels (the left panel of \autoref{fig:z4}b); instead, as seen in the right panel, their primary effect is on \textit{consonant} nasality. These results suggest that ciwGAN encodes English nasal vowels as an non-contrastive phonetic feature which co-occurs with nasal consonants, a phonological feature.

\begin{figure*}[htbp]
	\centering
	\subfloat[English like - $z$60 \& $z$68]{\includegraphics[width=0.49\linewidth,trim=50 34 50 50,clip]{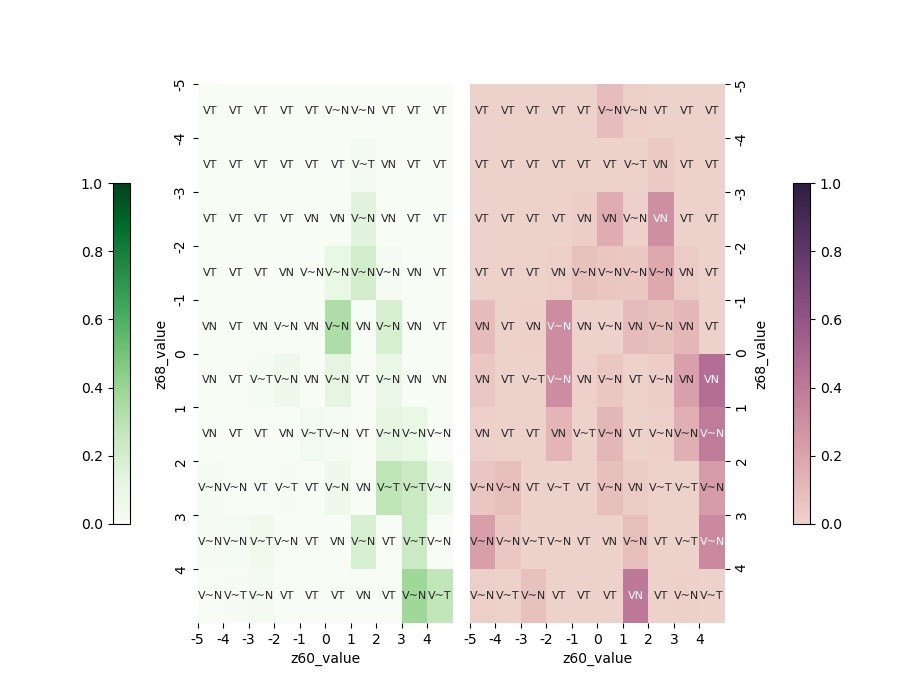}\label{fig:z68}}
	\hfill
	\subfloat[English like - $z$60 \& $z$71]{\includegraphics[width=0.49\linewidth,trim=50 34 50 50,clip]{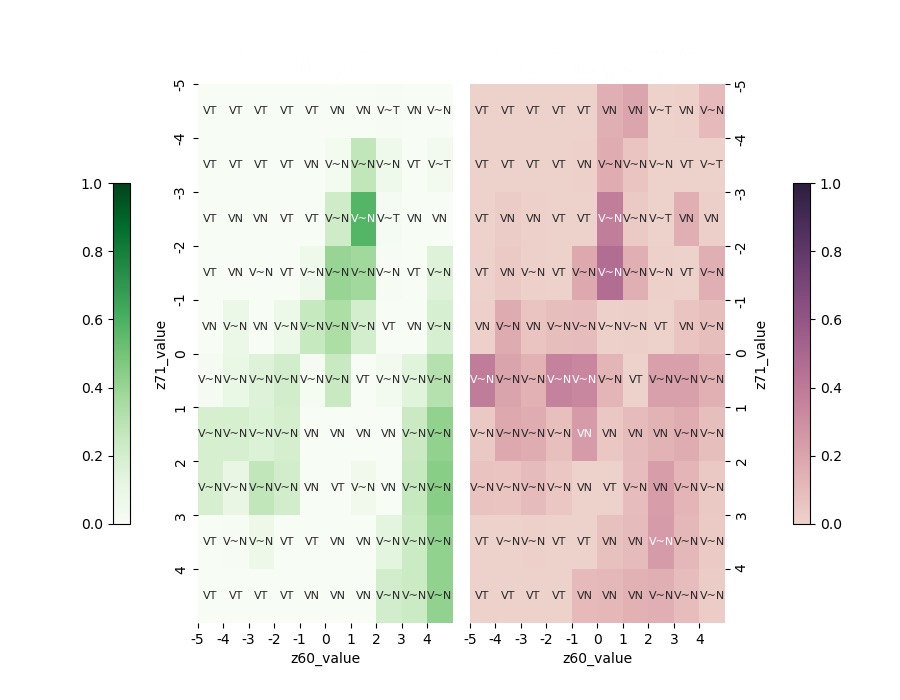}\label{fig:z71}}
    \vspace{-0.2cm}
	\caption{English-like experiment results: \autoref{fig:z68} shows the generated audios with concurrent manipulation of latent variables $z$60 and $z$68 (x axis: $z$60 \& y axis: $z$68); \autoref{fig:z71} shows the generated audios with concurrent manipulation of latent variables $z$71 and $z$60 (x axis: $z$71 \& y axis: $z$60) Green heatmap indicates the detected English nasal vowel; Red heatmap indicates the detected English nasal consonant.} 
    \vspace{-0.5cm}
\end{figure*}

\subsection{French Experiment}
\label{ssec:frn_experiment}
The networks learn to generate speech-like sequences (VT, VN, \H{V}T, \H{V}N) that are similar to training data as well as the distribution of nasalized vowels and oral vowels in French after 649 epochs' training. We perform the same analysis process as we had in English Experiment. Two latent variables ($z$4 and $z$37) are also found to be closely relative to French nasal consonants. Different from English, two latent variables ($z$88 and $z$91) show independent control of French nasal vowels. 

Manipulating these pairs of latent variables concurrently shows some interaction of latent variables in controlling nasal vowels and nasal consonants. In \autoref{fig:z37}, although $z$4 show primary controls of nasal consonants, as nasal consonants tend to presence in the generated outputs when $z$4 is positive, some interaction effects of $z$4 and $z$37 are found near the bottom right of the right panel.
In \autoref{fig:z88}, $z$88 and $z$91 demonstrates interactive effects on the nasal vowels: when $z$88 \textgreater 0 and $z$91\textless 0, the Generator tends to output nasal vowels. Most importantly, the variables tested in \autoref{fig:z88}a control nasal consonants while the ones in \autoref{fig:z88}b control vowels--- unlike the English results, in which one set of variables controlled both. These results indicate that both French nasal vowels and nasal consonants are encoded as independent phonological features in ciwGAN and ciwGAN seems to apply some interactions between latent variables to control the presence of phonological features.
\subsection{Balanced Training Dataset Experiments}
\label{ssec:bln_experiment}
In previous two experiments,we found that ciwGAN can capture the contrastiveness of the phonological phenomenon in English and French with different learned representation. We are also interested to evaluate how would the frequencies of different syllable types in the training data affect the learned representations of ciwGAN. 
We conduct experiments on two artificially balanced datasets. For our English-like experiment, we have 5570 tokens of the VT, 5570 tokens of VN. For French-like experiment, as most French nasal vowels extracted from SIWIS tend to be /\H{o}/, we mitigate this bias by only include tokens with vowel /o/ for all syllable types in the training dataset: 1031 tokens of the oT, 1031 tokens of oN, 1031 tokens of \H{o}T, 1031 tokens of \H{o}N.

\paragraph{English-like Experiment}
In contrast to the natural English ciwGAN, where no latent variables are found to control nasal vowels, the Generator seems to encode vowels' nasality with latent variables ($z$60, $z$71), even though latent variable $z$60 is found to controls the both nasal consonants and nasal vowels. By manipulating $z$60 to [-5, 5],  we can decrease the proportion of nasality in both vowels and consonants and have nasal vowels and nasal consonants completely disappear in the generated data.

Interactive effects are found between $z$60 and $z$68 and between $z$60 and $z$71 in controlling nasal consonants and nasal vowels respectively, which is similar to the interactive correlations of latent variables we found in French experiment.
As illustrated in \autoref{fig:z68} and \autoref{fig:z71}, the ciwGAN tends to generate nasal consonants except when the values of $z$60 and $z$68 are both set to negative and ciwGAN will generate nasal vowels when $z$60 and $z$71 are non-negative. Despite the dependency between nasal vowels and nasal consonants is also found in English ciwGAN with balanced dataset: the Generator tends to produce nasal vowels following nasal consonants, ciwGANs can generate independent nasal vowels in some generated audio: there are some tokens carry \H{V}T in the generated audios.

\begin{figure*}[htbp]
	\centering
	\subfloat[French-like - $z$60 \& $z$71]{\includegraphics[width=0.49\linewidth,trim=50 34 50 50,clip]{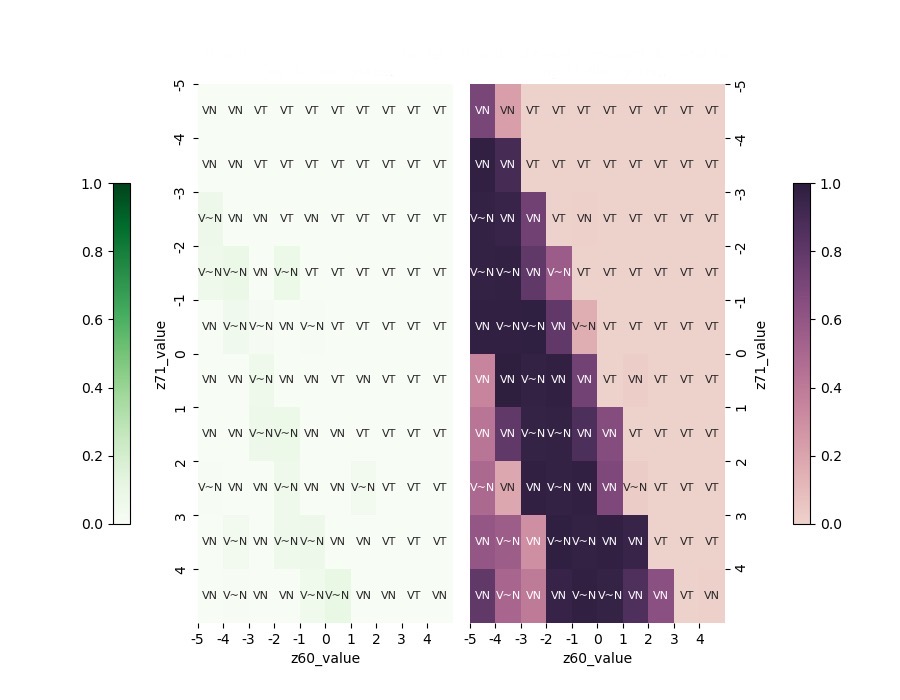}\label{fig:z6071}}
	\hfill
	\subfloat[French-like - $z$88 \& $z$16]{\includegraphics[width=0.49\linewidth,trim=50 34 50 50,clip]{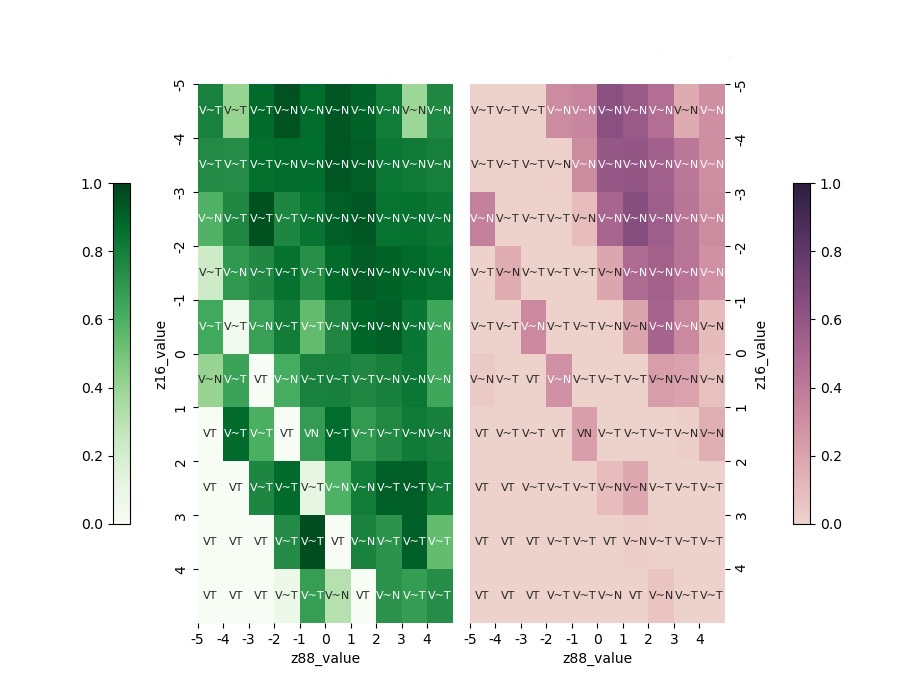}\label{fig:z8816}}
    \vspace{-0.2cm}
	\caption{French-like experiment results: \autoref{fig:z6071} shows the generated audios with concurrent manipulation of latent variables $z$60 and $z$71 (x axis: $z$60 \& y axis: $z$71); \autoref{fig:z8816} shows the generated audios with concurrent manipulation of latent variables $z$88 and $z$16 (x axis: $z$88 \& y axis: $z$16) Green heatmap indicates the detected French nasal vowel; Red heatmap indicates the detected French nasal consonant.} 
    \vspace{-0.5cm}
\end{figure*}
\paragraph{French-like Experiment}
With balanced dataset, we can still find latent variables that only control nasal consonants. As shown in \autoref{fig:z68} nasal consonants can be produced independently when $z$60 \textless 0 and $z$71 \textgreater 0.
Interactive effects of latent variables are also found on both nasal vowels and nasal codas. ciwGAN tend to generate nasal vowels when $z$16\textgreater 0 and $z$88 \textless 0, as in \autoref{fig:z71}. However, different from the model trained on natural French dataset, we cannot find latent variables that only control French nasal vowels. When $z$16 is set to a positive value and $z$88 is set to be negative, the generated audios on the top right of the \autoref{fig:z71}, are detected to have both nasal vowels and nasal consonants.

The phenomenon that interactive effects occurs in ciwGAN with balanced English dataset matches with the finding in French experiment and French-like experiment, which suggests that ciwGAN develops similar learned representations between the two languages with balanced datasets. Besides, no latent variables can only control French nasal vowels in French-like experiment, which is similar to the results in English-like experiments, but different from French experiment.
\section{Conclusion}

Our results qualify \newcite{begus2020generative}'s claim that GANs can learn clearly interpretable representational systems in which single latent variables correspond to identifiable phonological features. While we do find this in the English experiment, we do not find it in the French experiment, English-like experiment and French-like experiment. This suggests that both the frequencies with which different syllable types in the data occur, and the contrastiveness of the phonological phenomenon, may affect whether the learned representation is simple or distributed across many variables. Moreover, as the learned representations in ciwGANs involve featural conjunction, this counters \newcite{begus2020generative}'s claim of ciwGANs having an independent dimension for every phonological feature. In future work, understanding more complicated feature interactions, we plan to use eigendecomposition or other methods which can more easily represent higher-order interactions between features. However, our current methods are still informative about the learned representations, since the regression analyses show that only a few of the learned features are critical to representing nasality.

On the other hand, we do find that GANs clearly distinguish between the contrastive and non-contrastive status of vowel nasality in English and French. This supports \newcite{begus2020generative}'s higher-level claim that GANs are good phonological learners by testing it in a more controlled setting in which the same feature is compared across languages.

While artificially balancing the frequencies of syllable types in the training data does not erase the difference between English and French, we do observe that the learned representations are more similar between the two, and that the GANs learning from English data begins to be able to generate some \H{V}T syllables, although with low frequency. This aligns with a widespread theory for the origin of contrastive nasality in languages like French. Changing the patterns’ frequency will change the feature systems in languages.

Our results highlight the difficulty of learning featural phonological representations from acoustic data, as well as the interpretational difficulties of detecting such representations once learned. We believe that the question of which architectures successfully acquire these systems is still open--- more work needs to be done on larger pretrained models to determine which, if any, of these generalizations they encode. More careful comparisons between smaller-scale systems can also shed light on how well they distinguish between completely predictable (allophonic) distributional properties of segments due to phonotactic constraints, and statistical regularities due to the lexicon or morphology.

On the other hand, the observed difficulty of learning these generalizations lends support to theories of phonological change in which mistakes in acquisition lead to the expansion or restructuring of a feature inventory \cite{foulkes2013first}. By looking at historical corpus of old French, we can observe how the lexicon evolves over time changing the frequency of different vowel-consonant combinations. The fact that changes in frequency result in this kind of change for our model is evidence that this mechanism is plausible, and offers a route to testing its explanatory power for specific historical hypotheses in the future.

Although the long-term goal of this research is understanding how phonological representation learning works for a variety of models and phenomena, we believe it is necessary to start small, with the treatment of one particular phenomenon. In text linguistics, there are now established benchmarks for understanding linguistic representation in language models, for example, The Benchmark of Linguistic Minimal Pairs (BLiMP) \cite{warstadt2020blimp}, but in speech linguistics, we are lagging behind. Even doing studies of an individual phenomenon requires identifying a phonological phenomenon, extracting and labeling a corpus and conducting a study of the model’s learning behavior. A diverse and comprehensive benchmark dataset for studying phonological learning (beyond phoneme segmentation and categorization) would be an exciting goal for future work.

\section{Acknowledgements}
We thank the Phonies group at OSU Linguistics Department for helpful discussion, especially Cynthia Clopper and Becca Morley. We also thank Ga\v{s}per Begu\v{s} for sharing the training dataset used in \cite{begus2022interpreting}

\section{Limitations}

The study of language model in their alignment to linguistic theories are interdisciplinary and hence usually hard to find explicit connection between language model and theories. In this paper we claim that a generative model, ciwGAN, can model both phonetic and phonology features. However, the two features are learned by two ciwGAN instances from disjoint training data sets. Our finding couldn't support or deny the following statements that are of researchers' concern:

\begin{enumerate}
    \item Generic GAN model can learn phonology features like ciwGAN.
    \item CiwGAN can model phonetic and phonology features simultaneously from a single dataset.
\end{enumerate}


\bibliographystyle{acl_natbib}
\bibliography{acl2023jingyichen}



\newpage
\appendix

\section{Manipulation Effects on Nasal Consonant}
\begin{figure}[ht]
	\centering
	\includegraphics[width=1\linewidth]{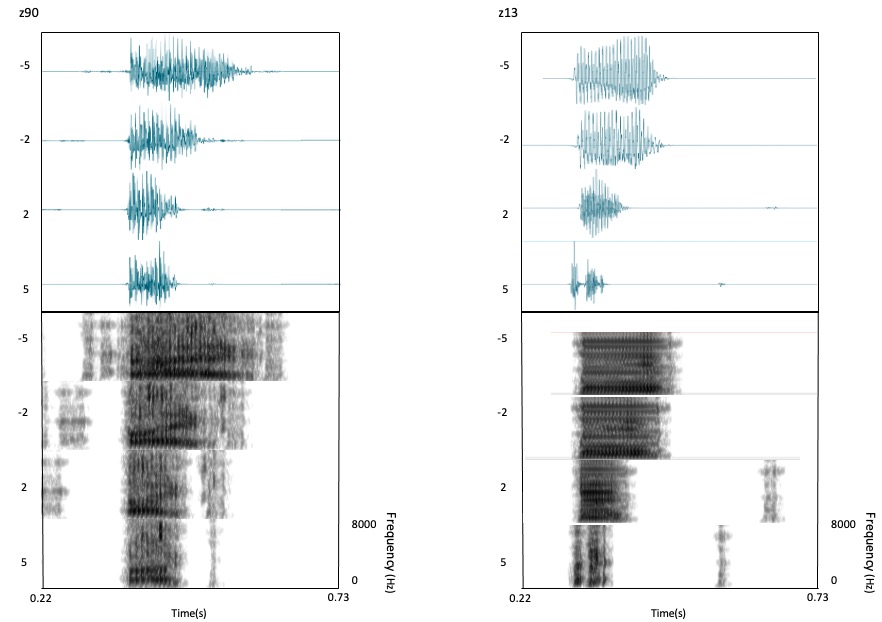}
    \vspace{-0.7cm}
	\caption{Waveforms and spectrograms (0-8000 Hz) of generated audio with $z$90 variable manipulated (left); Waveforms and spectrograms (0-8000 Hz) of generated audio with $z$13 variable manipulated (right)}
	\label{fig:_05_spectrom_praat_z90z13}
\end{figure}

\autoref{fig:_05_spectrom_praat_z90z13} illustrates the manipulation effects of $z$13 and $z$90 on nasal consonant.

\section{Example Lexical Items of French and English}
\begin{tabular}{ |p{1.1cm}||p{1.7cm}|p{1.4cm}|p{1.1cm}|p{1.5cm}|  }
 \hline
 \multicolumn{5}{|c|}{French Lexical Items from dataset} \\
 \hline
 Syllable types& Orthography & IPA & Gloss &Extracted part \\
 \hline
 CVT& pote & /p\textipa{O}t/ & "buddy" &/\textipa{O}t/\\
 CVN&   bon ami  & /b\textipa{O}nami/ & "good friend"   &/\textipa{O}n/\\
 C\H{V}T& ponte   &/p\H{\textipa{O}}t/&  "clutch" &/\H{\textipa{O}}t/ \\
 C\H{V}N & mon ami & /m\H{\textipa{O}}nami/ & "my friend" & /\H{\textipa{O}n}/ \\
 \hline
\end{tabular}

\begin{tabular}{ |p{1.5cm}||p{2cm}|p{1.5cm}|p{1.5cm}|  }
 \hline
 \multicolumn{4}{|c|}{English Lexical Items from dataset} \\
 \hline
 Syllable types& Orthography & IPA & Extracted part \\
 \hline
 CVT&  bad   &/b\ae d/&  /\ae d/\\
 CVN&   ban  & /b\ae n/   &/\ae n/\\
 
 \hline
\end{tabular}

\section{Model Parameters and Source Code}
\label{sec:appendix}
WaveGAN parameters and source code are provided in 
\url{https://github.com/DeliJingyiC/wavegan_phonology.git}

\clearpage
\section{Linear Regression Analysis}
\label{sec:appendix}
In section \ref{sec:experiment}, we have linear regression analysis
to identify latent variables that correlate to nasal features. The values of 100 latent variables in ciwGAN's latent space is analyzed and 7 latent variables that have the highest chi-square scores are considered to have a strongly correlation to nasality.

\begin{figure}[htbp]
	\centering
	\rotatebox{0}{\includegraphics[width=2\linewidth]{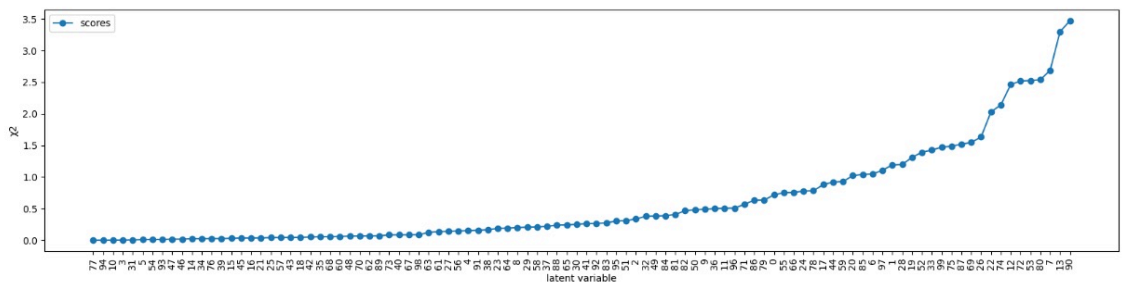}}
	\caption{Linear regression analysis of the nasality and the corresponding latent variables z. Y axis is Chi-square scores for 97 latent variable z and X axis is latent variables z.}
	\label{fig:chi}
\end{figure}

\end{document}